\documentclass[conference]{IEEEtran}
\IEEEoverridecommandlockouts
\usepackage{lipsum}
\usepackage{cite}
\usepackage{amsmath,amssymb,amsfonts, amsthm}
\usepackage{algorithmic, algorithm}
\usepackage{graphicx}
\usepackage{textcomp}
\usepackage{xcolor}

\usepackage{booktabs}
\usepackage{multirow}
\usepackage{enumitem, array}
\usepackage{caption}
\usepackage{subcaption}
\usepackage[export]{adjustbox}
\usepackage{hyperref}
\usepackage{gensymb}

\def\T{{\bf T}}
\def\X{{\bf X}}
\def\t{{\bf t}}
\def\x{{\bf x}}
\def\f{{\bf f}}
\def\y{{\bf y}}

\def\BibTeX{{\rm B\kern-.05em{\sc i\kern-.025em b}\kern-.08em
    T\kern-.1667em\lower.7ex\hbox{E}\kern-.125emX}}

\newtheorem{thm}{Theorem}[section]
\newtheorem{lem}[thm]{Lemma}

\theoremstyle{definition}

\begin{document}

\title{Mathematical model of printing-imaging channel for blind detection of fake copy detection patterns

\thanks{S. Voloshynovskiy is a corresponding author.}
\thanks{This research was partially funded by the Swiss National Science Foundation SNF No. 200021\_182063.}
}
 
\author{\IEEEauthorblockN{Joakim Tutt, Olga Taran, Roman Chaban, Brian Pulfer, Yury Belousov, Taras Holotyak and Slava Voloshynovskiy}
\IEEEauthorblockA{Department of Computer Science, University of Geneva, Switzerland \\
\{joakim.tutt, olga.taran, roman.chaban, brian.pulfer, yury.belousov, taras.holotyak, svolos\}@unige.ch}
}

\maketitle

\begin{abstract}
Nowadays, copy detection patterns (CDP) appear as a very promising anti-counterfeiting technology for physical object protection. However, the advent of deep learning as a powerful attacking tool has shown that the general authentication schemes are unable to compete and fail against such attacks. In this paper, we propose a new mathematical model of printing-imaging channel for the authentication of CDP together with a new detection scheme based on it. The results show that even deep learning created copy fakes unknown at the training stage can be reliably authenticated based on the proposed approach and using only digital references of CDP during authentication.
\end{abstract}

\begin{IEEEkeywords}
copy detection patterns, authentication, predictor channel, one-class classification, deep learning fakes.
\end{IEEEkeywords}

\section{Introduction}

Nowadays, counterfeiting and piracy are among the main challenges for modern economy. Existing methods of anti-counterfeiting are very diverse, ranging from watermarking techniques, special inking, holograms, electronic IDs, etc. The drawbacks of these technologies are that they can be expensive, often proprietary, and usually, authentication is performed in a non-digital way.

\begin{figure}[htbp]
\centerline{\includegraphics[scale=.87, trim={0cm, 0cm, 0cm, -0.13cm}]{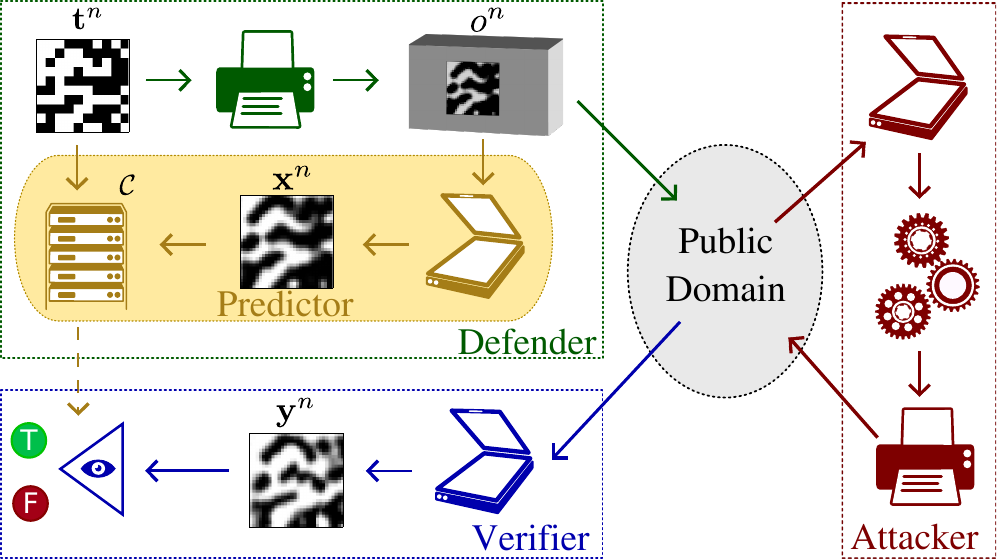}}
\caption{Illustration of the PI channel seen as a 3-player game. The Defender (in green) generates and prints template $\t^n$ on an object package $o^n$ and sends it to the public domain. The Attacker (in red) can use $o^n$ to create a counterfeited version $c^n$ of it. Finally, the Verifier (in blue) scans the object package and authentifies the probe $\y^n$ to decide whether it is an original package or a counterfeit. The novelty of our model is adding a predictor channel (in yellow) based on a codebook $\mathcal{C}$ which is trained by the Defender and used by the Verifier to enhance the classification results at the authentication stage.}
\label{fig:defender_attacker_scheme}
\end{figure}

A newly promising emerged field in digital anti-counterfeiting technologies is the usage of Printing Unclonable Features (PUF) which are based on intrinsic forensic uncloneable features of physical objects, such as randomness of ink blots or paper micro-structures \cite{zhu2003print, adams2011study, voloshynovskiy2012towards}. Another technology is the Copy Detection Patterns (CDP) \cite{picard2004digital} which are random binary patterns of high entropy that are difficult to clone, such as very small sized QR codes. The advantages of CDP, in comparison to other technologies, are that they are cheap, easily integrable with a product into a structure of QR-code and digitally readable \cite{picard2021counterfeit}. They are also easy to integrate in a track-and-trace distribution framework. The main challenge of this technology today is that, although being mainly robust to common copy attacks when simple decision rules are used based on the similarity to the reference template blueprint, it faces significant difficulties with the advanced machine-learning (ML) copy attacks. The possibility to use powerful deep classifiers in two-class classification allows one to reliably distinguish original CDPs from fakes, if the fakes used at testing time match the statistics of those used during training. However, in the case of mismatches, the method fails to distinguish original and fakes~\cite{taran2022mobile}. In practice, the situation is further complicated by several factors:

\begin{itemize}
    \item the high deviations in printing and imaging leading to large intra-class variabilities;
    \item ML attacks that are able to produce blueprint estimations with an accuracy score as high as $94\%$ \cite{chaban2021machine};
    \item the natural lack of exact prior knowledge for the authenticator about the fakes in field. Fakes can be produced in multiple ways and it is unknown which fake will be used at the attacking time;
    \item the absence of a reliable model of printing-imaging channel that complicates the design of optimal authentication rules.
\end{itemize}

Therefore, there is a critical need in a one-class (OC) authentication scheme able to operate in the generalized setup of the above printing-imaging channel without prior knowledge of the fakes. In this paper we adress these problems by:

\begin{itemize}
    \item providing a new stochastic model describing the defender Printing-Imaging (PI) channel;
    \item proposing a new method of authentication based on the PI model able to perform authentication in the OC-classifier setup, i.e., under complete ignorance about the actual fakes;
    \item validating the proposed approach on a real dataset of CDPs of originals and ML-based fakes based on codes designed with $1 \times 1$ symbol and produced on two industrial printers;
    \item comparing the proposed method with traditional authentication techniques.
\end{itemize}

The paper is organized as follows. Section \ref{section:problem_formulation} introduces the problem formulation and presents a stochastic model of PI channel for the defender that forms the basis of the OC-classification framework. Section \ref{section:oc_classification_algo} presents the algorithm of OC-classification for CDPs in two variations. Section \ref{section:results} presents the results of performance and comparison with standard metrics on the same dataset. Finally, Section \ref{section:conclusion} concludes the paper and discusses possible extensions and perspectives. All mathematical notations used in the paper can be found in Table \ref{table:notation_table}.

\section{Problem formulation}
\label{section:problem_formulation}

\subsection{The printing-authentication scheme}

The production of an anti-counterfeit technology using CDP is best described as a 3-player game with a Defender, an Attacker and a Verifier as shown in Fig. \ref{fig:defender_attacker_scheme}.

The Defender protects his brand by using a family of digital CDP blueprints $\{ \t^n \}_{n=1}^N$ stored in the form of a binary matrix $\t^n$, which is then printed on the object package $o^n$ and sent to the public domain.
The Attacker has access to the printed version of the CDP and may use it to create a counterfeit $c^n$, through the process of scanning, post-processing and reprinting (see \cite{chaban2021machine, khermaza2021can, yadav2019estimation, taran2019clonability} for investigations of attacking techniques).
At the authentication stage, the Verifier receives an unindentified package (either $o^n$ or $c^n$) from which a digital image $\y^n$ is acquired, using any device such as a scanner or a mobile phone. We denote $\x^n$ the code acquired from $o^n$ and $\f^n$ the code acquired from $c^n$.
An authentication is then performed based on the probe $\y^n$, which might be either $\x^n$ or $\f^n$, and on the reference template $\t^n$.

\subsection{Authentication techniques}

The algorithms used for authentication evolved a lot in the last few years. At first, CDP were designed with an idea to be resistant to simple scanning \& reprinting attacks~\cite{picard2004digital}. Due to the dot gain effect of printers, a portion of the information stored in the template blueprint $\t$ is lost in the probe $\y$ through the process of printing and scanning. Various ways to measure the information loss have been proposed which can be formalized with different types of metrics:

\newcommand{\ra}[1]{\renewcommand{\arraystretch}{#1}}
\begin{table}
\caption{Mathematical notations used in the paper.}
\ra{1.3}
\begin{tabular}{rcl}
\toprule
& Mathematical notation & Meaning \\
\midrule
& $\t$              & binary digital template \\
& $\x$              & digital original printed from $\mathbf{t}$ \\
CDPs & $\f$         & digital fake version of $\mathbf{t}$ \\
& $\y$              & probe representing either $\mathbf{x}$ or $\mathbf{f}$ \\
& $\tilde{\t}$      & digital template estimated from $\mathbf{y}$ \\

\midrule
& $\T$                           & binary random matrix for $\mathbf{t}$ \\
& $\X$                           & random matrix for $\mathbf{x}$ \\
& $\tilde{\T}$                   & binary random matrix for $\mathbf{\tilde{t}}$ \\
PI Model & $p \in [0,1]$        & probability of black symbol in $\T$ \\
& $\omega \in \Omega$           & set of all neighbourhoods \\
& $P(\omega)$                   & positive probability at $\omega$ \\
& $P_b(\omega)$                 & probability of bit-flipping at $\omega$ \\
& $\mathcal{C}$                 & codebook of probabilities \\
\midrule

& $n = 1, ...., N$              & index within the dataset \\
& $(i,j)$ or $(r,s)$            & coordinates of pixels in $\mathbf{t}$ \\
Numbers & $L \times L$       & size of $\mathbf{t}$ \\
& $h = 1,3,5, ...$         & integer defining the size of $\omega$ \\
& $k = 1,2,3, ...$              & magnification factor from $\mathbf{t}$ to $\mathbf{x}$ \\
\bottomrule
\end{tabular}
\label{table:notation_table}
\end{table}

\begin{figure}[htbp]
\centerline{
\includegraphics[scale=.45, trim={3.cm, 2.cm, 2.5cm, 1.2cm}, clip]{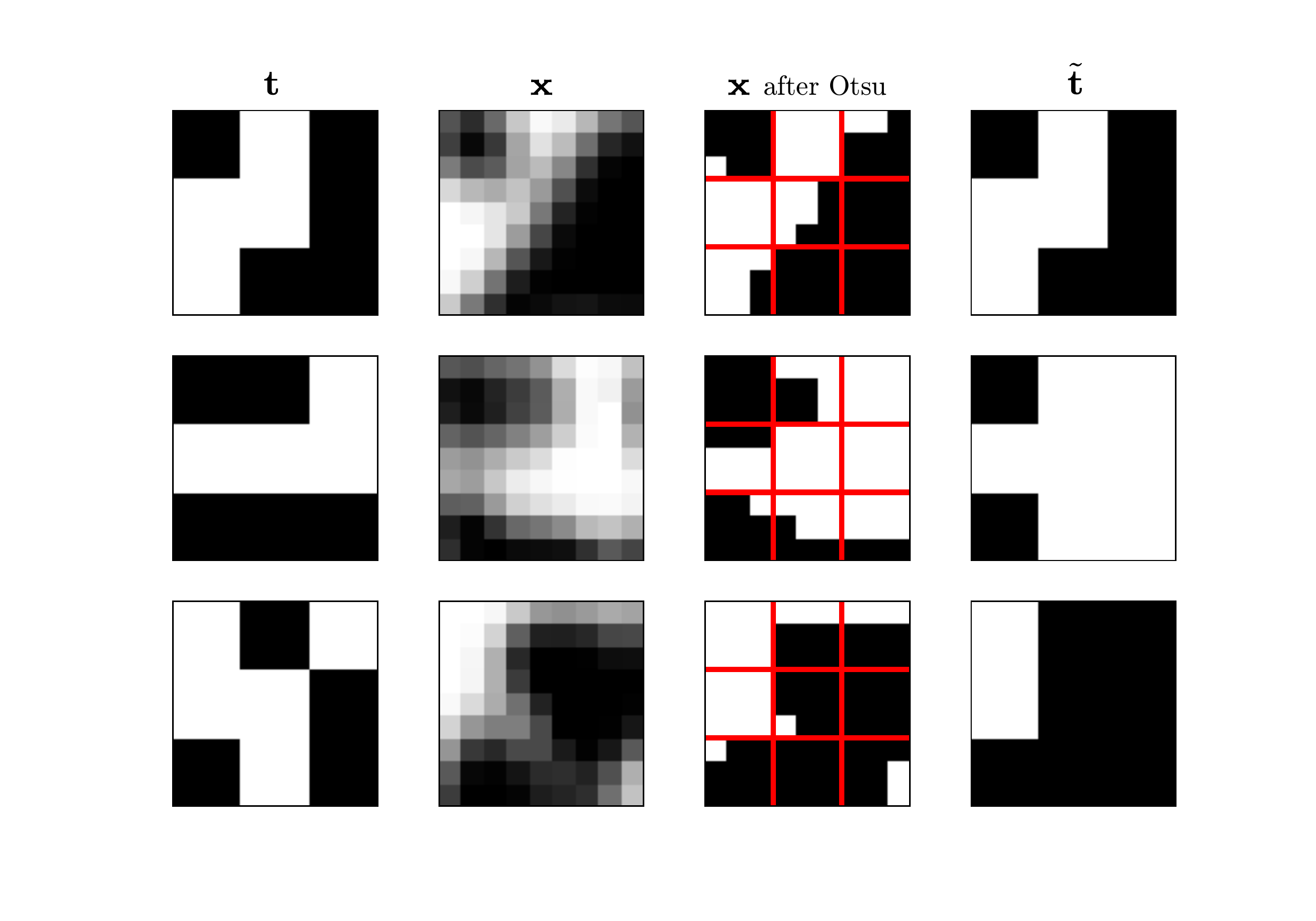}
}

\caption{Otsu's binarization technique followed by majority voting. The first column shows different neighbourhoods $\omega$, the second column the printed originals $\x$, the third column $\x$ after Otsu's binarization and the fourth column, the estimated template $\mathbf{\tilde{t}}$ after majority voting. Red lines highlight the $3 \times 3$ patches in $\x$ corresponding to one symbol in $\t$. Different types of distorsions are illustrated leading to estimation errors in $\tilde{\t}$.}
\label{fig:otsu_majority_voting}
\end{figure}

\begin{enumerate}
    \item $\ell_1$- or $\ell_2$-distance between the probe $\y$ and the template $\t$;
    \item Pearson correlation between $\t$ and $\y$;
    \item Hamming distance between the template $\t$ and an estimation $\tilde{\t}$ of the template, based on the probe $\y$. A very common way to perform the estimation $\tilde{\t}$ is to use Otsu's binarization algorithm and then a majority voting for each symbol. Fig. \ref{fig:otsu_majority_voting} on the next page illustrates this technique.
\end{enumerate}

Nowadays, new techniques emerge with the use of machine learning, allowing one to train deep classifiers \cite{taran2022mobile, cui2020new} and deep binarization techniques \cite{chaban2021machine, khermaza2021can, yadav2019estimation, taran2019clonability}. Although showing very promising results, these new algorithms act as black boxes and thus lack interpretability, which is paramount when working on reliability questions and security-critical applications such as the protection of pharmaceutical products.

\subsection{Stochastic model of Printing-Imaging channel}

The PI channel can be described mathematically as a Markov Chain $\T \rightarrow \X \rightarrow \tilde{\T}$, where:

\begin{itemize}
    \item $\T$ is a random binary matrix of size $L \times L$ sampled from i.i.d. Bernoulli distribution: $T_{ij} \sim Bern(p)$, $p \in [0,1]$ is the probability of black symbol;
    \item $\X$ is a random matrix of size $kL \times kL$, $X_{ij} \in [0,1]$ for some magnification factor\footnote{The magnification factor is related to the resolution of enrollment equipment. Nowadays, with modern scanners and mobile phones, $k \geq 1$.}
     $k = 1, 2, 3, ...$;
    \item $\tilde{\T}$ is a random binary matrix of size $L \times L$.
\end{itemize}

In reality, when we pass a template $\t$ through the PI channel, some distorsions occur in $\x$ due to the dot-gain effect and printing-related natural randomness. Thus, when we try to estimate $\tilde{\t}$ from $\x$, we end up with some errors, dependant on the printer, type of paper, acquisition device and chosen estimator. In this paper, we are mostly interested in understanding the probability distribution $\mathbb{P}(\tilde{\T} | \T)$, which we believe to be highly correlated with the particular choices of print-acquire-estimate system and is central when trying to estimate information loss.

In \cite{voloshynovskiy2016physical}, the authors model this probability distribution as a Binary Symmetric Channel (BSC). This model assumes that each symbol $T_{ij}$ in $\T$ has a certain probability $P_b$ of bit-flipping, independently of its location $(i,j)$. We conjecture that the BSC model is too simple to capture the random behaviour of printing, as it does not take into account the local dependency of neighbouring sites and rather learns an average probability of bit-error across the whole template. Another related model with multilevel symbols has been studied in \cite{villan2006multilevel}. Inspired by the BSC model, we introduce a new stochastic model with three key assumptions:

\begin{enumerate}
    \item \emph{Markovianity:} the posterior probability at a particular symbol location $(i,j)$ only depends on the local neighbourhood $\omega_{ij}$ surrounding it:
    \begin{equation}
        \mathbb{P}(\tilde{T}_{ij} | \T) = \mathbb{P}(\tilde{T}_{ij} | \omega_{ij}),
        \label{eq:locality_axiom}
    \end{equation}
    
    where $\omega_{ij}$ is a small neighbourhood surrounding symbol $T_{ij}$, typically a square matrix centered around $(i,j)$: 
    \[
    \omega_{ij} = \{ T_{i \pm a, j \pm b} | 0 \leq a,b < h/2 \},
    \]

    where $h = 1, 3, 5, ...$ is fixed by the model and defines the size of the neighbourhood.
    
    \item \emph{Stationarity:} the posterior probability does not depend on the location inside the image. Similar patterns in $\T$ lead to similar probability values\footnote{
    The printing and scanning process introduces a lot of variability. The goal of the model is not to learn the fingerprint of a particular realization but rather measure the average variability for each neighbourhood and to take advantage of this knowledge. \eqref{eq:stationarity_axiom} should be read as an equality in distribution, allowing every realisation of $\tilde{T}_{i,j}$ to be different while still following a common law, independent of the location $(i,j)$.}:
    \begin{equation}
        \mathbb{P}(\tilde{T}_{ij} | \omega_{ij}) = \mathbb{P}(\tilde{T}_{rs} | \omega_{rs}), \mbox{ if } \omega_{ij} = \omega_{rs}.
        \label{eq:stationarity_axiom}
    \end{equation}
    
    \item \emph{Posterior independance:} the joint posterior probability factorizes as:
    \begin{equation}
        \mathbb{P}(\tilde{\T} | \T) = \prod_{i,j} \mathbb{P}(\tilde{T}_{ij} | \T ).
        \label{eq:posterior_independance_axiom}
    \end{equation}
\end{enumerate}

With assumptions \eqref{eq:locality_axiom} and \eqref{eq:stationarity_axiom}, one can easily prove the expectation formula for the posterior distribution:
\begin{equation}
\mathbb{P}(\tilde{T}_{ij} | \omega_{ij}) = \mathbb{E}_{r,s: \omega_{rs} = \omega_{ij}} [ \mathbb{P}(\tilde{T}_{rs} | \omega_{rs})].
\end{equation}
This formula is a key to the proposed authentication scheme as it can be estimated directly using Monte-Carlo method from a training dataset. For each type of neighbourhood $\omega \in \Omega$ (there can be at most $2^{h^2}$), we learn the probability distribution which is highly correlated with the PI channel on which it was trained. Two measures associated with this distribution are the posterior probability of bit-flipping $P_b(\omega)$ and the positive posterior probability $P(\omega)$, which we define as:
\begin{IEEEeqnarray}{rCl}
    P_b(\omega_{ij}) &:=& \mathbb{P}(\tilde{T}_{ij} \neq T_{ij} | \omega_{ij}), \\
    P(\omega_{ij}) &:=& \mathbb{P}(\tilde{T}_{ij} = 1 | \omega_{ij}).
\end{IEEEeqnarray}

We can thus create a codebook in which we store all these different probability values for each type of neighbourhood and use them as references in the authentication scheme.

\subsection{Metric in the PI channel}

The introduced PI channel gives us a theoretical tool to better understanding the process of printing and acquisition of CDP. In this subsection, we show that this model comes with a very natural metric that can be easily implemented and used for authentication.

\begin{lem}
In the PI channel model, the posterior log-likelihood can be computed as:
\begin{equation}
\log \mathbb{P}(\tilde{\T} = \tilde{\t} | \T) = \sum_{i,j} \log (1 - | \tilde{t}_{ij} - P(\omega_{ij}) |).
\label{eq:lls_formula}
\end{equation}
\end{lem}

\begin{proof}
The proof relies on two steps. The first one is to use conditional independence of the symbols in $\tilde{\T}$ given $\T$ and Markovianity:
\begin{IEEEeqnarray*}{rCl}
    \log \mathbb{P}(\tilde{\T} = \tilde{\t} | \T) &=& \sum_{i,j} \log \mathbb{P}(\tilde{T}_{ij} = \tilde{t}_{ij} | \T) \\
        &=& \sum_{i,j} \log \mathbb{P}(\tilde{T}_{ij} = \tilde{t}_{ij} | \omega_{ij}).
\end{IEEEeqnarray*}

The second step is then a simple case study for $\tilde{t}_{ij} \in \{0,1\}$:

\begin{equation*}
\mathbb{P}(\tilde{T}_{ij} = \tilde{t}_{ij} | \omega_{ij}) = 1 - | \tilde{t}_{ij} - P(\omega_{ij}) |.
\end{equation*}
\end{proof}

\section{one-class CDP classification algorithms}
\label{section:oc_classification_algo}
The core idea of building an authentication system based on the PI channel model is to introduce \emph{the predictor channel}, which is trained using both digital templates $\t$ and acquired originals $\x$ and to learn a codebook $\mathcal{C}$ of probabilities for each neighbourhood $\omega \in \Omega$.

To train the predictor, we create two dictionaries $\mathbb{D}$ and $\mathbb{D}_b$ whose keys are the different types of neighbourhoods. For each $\omega_{ij} \in \Omega$, $\mathbb{D}[\omega_{ij}]$ lists the corresponding values of symbol $\tilde{t}_{ij}$ and $\mathbb{D}_b[\omega_{ij}]$ lists the boolean values $(\tilde{t}_{ij} \neq t_{ij})$. Finally, we compute the codebook $\mathcal{C}$, which is a database storing the statistics $P(\omega)$ and $P_b(\omega)$ for each type of neighbourhood $\omega$. A pseudo-code is given in Algorithm \ref{predictor_algo}.

\subsection{The likelihood score model}
\label{ssec:lls_model}
The first authentication scheme is a direct implementation of~\eqref{eq:lls_formula}. It starts by learning the codebook $\mathcal{C}$, running Algorithm~\ref{predictor_algo} on the training set. For the authentication of a probe $\y$, we perform the following steps:

\begin{enumerate}
\item estimate $\tilde{\t}$ from the probe $\y$;
\item with the reference template $\t$, search the probability $P(\omega_{ij})$ in $\mathcal{C}$, for each neighbourhood $\omega_{ij}$ in $\t$;
\item compute the likelihood score of $\tilde{\t}$ applying \eqref{eq:lls_formula};
\item compare the score with a chosen threshold fixed on the validation set to decide whether $\y$ is original or fake.
\end{enumerate}

It should be pointed out here that symbols $t_{ij}$ located too close to the border of the template do not have a well-defined neighbourhood $\omega_{ij}$. We propose two solutions to address this problem:

\begin{itemize}
    \item the first solution is simply to ignore these symbols and run the model only on the symbols located in the inside of $\t$;
    \item another solution is to consider a white padding surrounding template $\t$ as this is the natural padding for $\x$ when printing CDP on white paper.
\end{itemize}

\subsection{The attention model}
\label{ssec:attention_model}
The attention model is similar in essence to the preceding model but differs in several ways. The idea here is to use the probability bit-error map $P_b(\omega_{ij})$ as a mask, only keeping symbols that have a low probability of bit-error on the training set. In this way, we remove all regions in $\y$ that are known to produce high error for original samples $\x$. Training is done similarly to the likelihood score model above. For the authentication, we do:

\begin{algorithm}
 \caption{Algorithm for predictor training}
 \begin{algorithmic}[1]
 \label{predictor_algo}
 \renewcommand{\algorithmicrequire}{\textbf{Input:}}
 \renewcommand{\algorithmicensure}{\textbf{Output:}}
 \REQUIRE training set $\{ (\mathbf{t}^n, \mathbf{x}^n) \}_{n=1}^N$
 \ENSURE  learned codebook $\mathcal{C} = (\omega, P(\omega), P_b(\omega))_{\omega \in \Omega} $
 \\ \textit{Initialisation}:
  \STATE create two dictionaries $\mathbb{D}$ and $\mathbb{D}_b$ with the set $\Omega$ as keys and empty lists as values.
  \FOR {$n = 1$ to $N$}
    \STATE estimate $\tilde{\t}^n$ from $\x^n$
    \FOR {symbol $t_{ij}^n$ in $\t^n$}
        \STATE extract neighourhood $\omega_{ij}^n$ in $\mathbf{t}^n$
        \STATE extract symbol $\tilde{t}_{ij}^n$ in $\tilde{\t}^n$
        \STATE append value $\tilde{t}_{ij}^n$ in dictionary $\mathbb{D}$ at key $\omega_{ij}^n$
        \STATE append boolean value $(\tilde{t}_{ij}^n \neq t_{ij}^n)$ in dictionary $\mathbb{D}_b$ at key $\omega_{ij}^n$
    \ENDFOR
  \ENDFOR
  
  \FOR {$\omega$ in $\Omega$}
    \STATE compute mean value: $P(\omega) = \mbox{mean} (\mathbb{D}[\omega])$
    \STATE compute mean value: $P_b(\omega) = \mbox{mean} ( \mathbb{D}_b[\omega] )$
    \STATE store the triple $(\omega, P(\omega), P_b(\omega))$
  \ENDFOR
 
 \RETURN codebook $\mathcal{C} = (\omega, P(\omega), P_b(\omega))_{\omega \in \Omega}$
 \end{algorithmic} 
 \end{algorithm}

\begin{enumerate}
\item for each neighbourhood $\omega_{ij}$ in $\t$, search the probability of bit-flipping $P_b(\omega_{ij})$ in the codebook;

\item define an attention mask $m_{ij} := (P_b(\omega_{ij}) < \mu)$ for some fixed threshold $\mu \in [0,1]$;

\item choose any standard metric that is computed pixel-wise such as mean squared error, Hamming distance or Pearson correlation. Note that some upsampling of $\t$ might be necessary for computation;

\item weight the chosen metric $d(\t,\y)$ by using the binary mask, upsampling it if needed:
\[
d^{m}(\t,\y) = \sum_{i,j} m_{ij} \cdot d(t_{ij}, y_{ij}).
\]
\end{enumerate}

\section{Results}
\label{section:results}

\subsection{Dataset choice}

For our experiments, we use the Indigo $1 \times 1$ base dataset, presented in \cite{chaban2021machine}.
It is constituted of 720 different templates~$\t$ printed with two different printers: HP~Indigo 5500 DS (HPI55) and HP Indigo 7600 DS (HPI76) at 812.8 dpi, which we refer to as $\x^{55}$ and $\x^{76}$.
It also includes ML-based fakes of four different types: $\f^{55/55}$, $\f^{76/55}$, $\f^{55/76}$ and $\f^{76/76}$ where fake $\f^{mm/nn}$ is obtained from $\x^{nn}$ by the process of deepnet-based binarization, printed using HPImm and rescanned.

In this work, we only concentrate on the templates with $50 \%$ density of black symbols.
The templates $\t$ have a size of $228 \times 228$ symbols while $\x$ and $\f$ have a size of $684 \times 684$, that is a magnification by a factor $k=3$. We fix the training set size to $50$ samples, validation set to $100$ samples and test set to $500$ samples.

\begin{table*}
\caption{Results in percent of the Area Under Curve (AUC) for each type of originals and fakes and various metrics. Best results for each type of fakes are highlighted. Average results are shown per printer and in total.}
\centering
\ra{1.3}
\begin{tabular}{@{}rlllllcllllllcl@{}}
\toprule
& \multicolumn{5}{c}{HPI55 originals $x^{55}$} & \phantom{abc} & \multicolumn{5}{c}{HPI76 originals $x^{76}$} & \phantom{c} & \\
\cmidrule{2-6} \cmidrule{8-12}
& $f^{55/55}$ & $f^{55/76}$ & $f^{76/55}$ & $f^{76/76}$ & Average
&& $f^{55/55}$ & $f^{55/76}$ & $f^{76/55}$ & $f^{76/76}$ & Average && Total \\ \midrule
$metrics$\\
LLS & $99.88$ & $99.89$ & $100$ & $100$ & $99.94$
&& $87.24$ & $85.69$ & $\mathbf{99.97}$ & $\mathbf{99.98}$ & $93.22$ && $96.58$\\
MSE & $59.60 $ & $59.05$ & $35.85$ & $27.77$ & $45.57$
&& $97.47 $ & $\mathbf{99.03}$ & $85.85$ & $82.12$ & $91.12$ && $68.34$\\
PCOR & $87.11 $ & $88.41$ & $94.75$ & $92.36$ & $90.66$
&& $88.97 $ & $90.49$ & $95.79$ & $93.88$ & $92.28$ && $91.47$\\
HAMM & $63.39$ & $63.82$ & $69.46$ & $61.12$ & $64.45$
&& $85.36$ & $86.45$ & $89.35$ & $83.14$ & $86.08$ && $75.26$\\
$masked$\\
M-LLS & $\mathbf{99.98}$ & $99.96$ & $\mathbf{100}$ & $\mathbf{100}$ & $99.99$
&& $\mathbf{99.29}$ & $98.97$ & $99.94$ & $99.84$ & $\mathbf{99.51}$ && $\mathbf{99.75}$\\
M-MSE & $99.97$ & $99.95$ & $100$ & $100$ & $99.98$
&& $97.04$ & $94.76$ & $99.94$ & $99.85$ & $97.9$ && $98.94$\\
M-PCOR & $96.30$ & $92.46$ & $99.35$ & $98.58$ & $96.67$
&& $97.11$ & $95.45$ & $98.42$ & $97.72$ & $97.17$ && $96.92$\\
M-HAMM & $99.98$ & $\mathbf{99.97}$ & $100$ & $100$ & $\mathbf{99.99}$
&& $99.22$ & $98.89$ & $99.94$ & $99.85$ & $99.48$ && $99.73$\\
\bottomrule
\end{tabular}
\label{table:AUC_all_metrics}
\end{table*}

\begin{figure*}[t]
\centerline{\includegraphics[scale=.55, trim={1.5cm, 1.5cm, 1.5cm, 1.5cm}, clip]{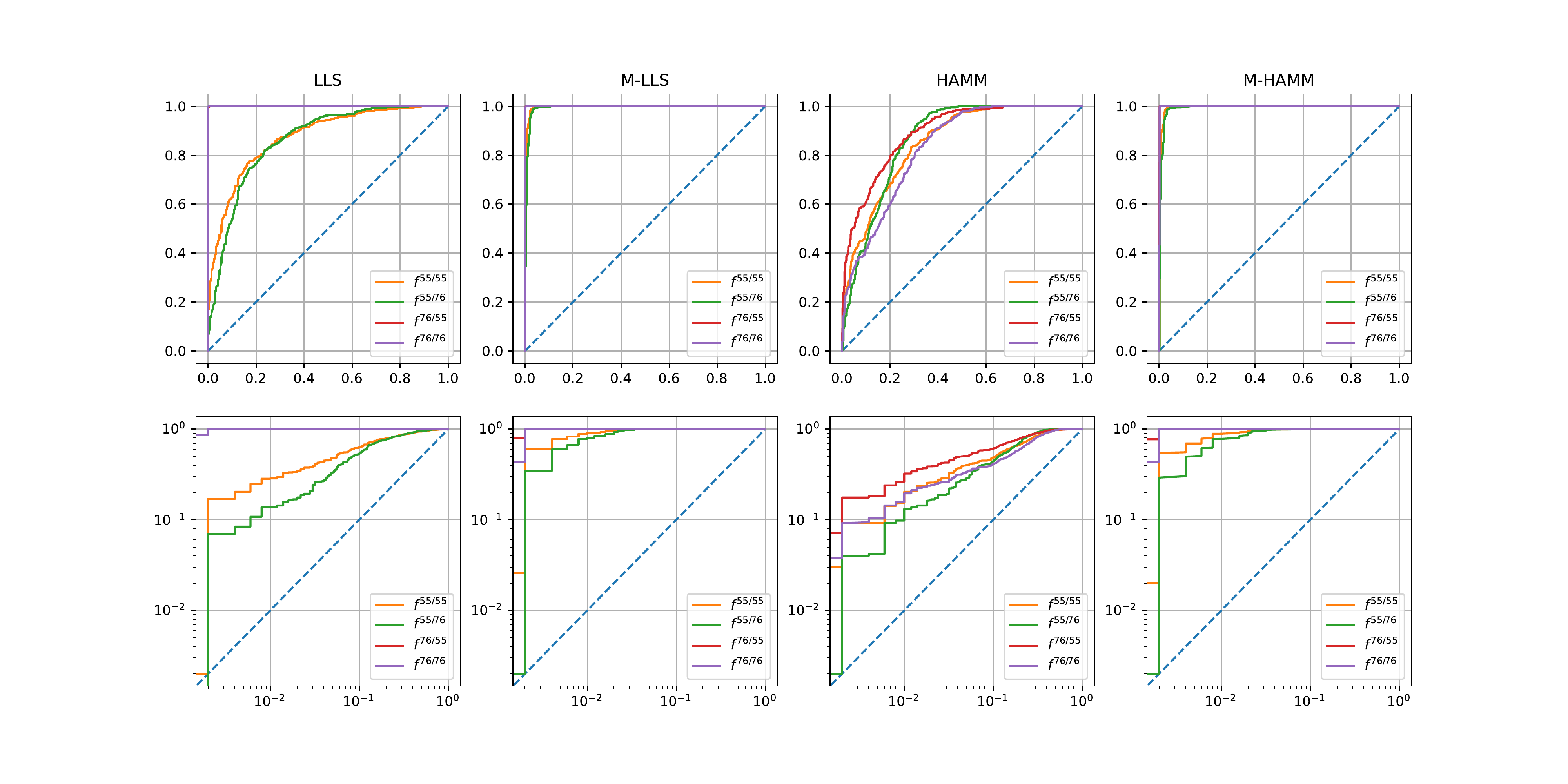}}
\caption{Visualisation of ROC curves for original $x^{76}$ and all four types of fakes. First column is the LLS metric, second column the masked LLS, third column the Hamming distance and fourth column the masked Hamming distance. The second row shows the same plots with logarithmic scale on both axes.}
\label{fig:roc_results}
\end{figure*}

\vspace{-.8mm}

\subsection{Predictor algorithm parameters}
\label{ssec:algo_param_choice}
In order to train the predictor, we fix a certain number of parameters.
The first one is the estimator $\x \rightarrow \tilde{\t}$. As we saw in Section \ref{section:problem_formulation}, there are many different approaches to it. We decide to use Otsu's algorithm for binarization followed by majority voting on each $3 \times 3$ patch corresponding to one symbol in $\t$.
We fix the size of neighbourhoods $\omega$ in $\t$ to be of size $3 \times 3$ for the following reasons:

\begin{itemize}
    \item This brings the total number of possible neighbourhoods down to $|\Omega| = 2^9 = 512$ which is small enough in comparison to the total number of neighbourhoods in a single template: $226^2 = 51'076$. We can thus expect to see every neighbourhood appear roughly $100$ times in each template.
    \item The printing process can produce some random deviations as we discussed in Section \ref{section:problem_formulation}, but these deviations are local in the sense that they only affect neighbouring symbols in most cases. Thus, $3 \times 3$ neighbourhoods are sufficient to capture them. See Fig. \ref{fig:otsu_majority_voting} for an illustration.
\end{itemize}

\subsection{Discussion}

To compare all different approaches in a unified way, we test both originals $\x^{55}$ and $\x^{76}$ separately against all four kind of ML fakes $\f^{55/55}$, $\f^{55/76}$, $\f^{76/55}$ and $\f^{76/76}$. The metrics that we use are:

\begin{itemize}
    \item the log-likelihood score (LLS) described in Section \ref{ssec:lls_model};
    \item the mean-squared-error (MSE) between $\y$ and $\t$;
    \item the Pearson correlation (PCOR) between $\y$ and $\t$;
    \item the Hamming distance (HAMM) between $\t$ and $\tilde{\t}$;
    \item the same four metrics mentioned above with a mask as described in Section \ref{ssec:attention_model}.
\end{itemize}

For each metric, we compute the associated ROC curves and report the AUC score. The AUC score is averaged over ten runs with randomization of training/testing set.
All results are summarized in Table \ref{table:AUC_all_metrics}.

A first observation at the results in Table \ref{table:AUC_all_metrics} shows that discriminating between originals and fakes is more accurate for $\x^{76}$ than for $\x^{55}$. In general, the results show that the metric LLS outperforms the other metrics. On average, M-LLS, its masked version, appears as the best metric with a very reliable AUC score on all types of fakes.

The masked metrics show a great improvement in AUC score over all their non-masked counterparts. This is further illustrated in Fig. \ref{fig:roc_results}, where we compare side-by-side masked and non-masked metrics for LLS and Hamming metrics.

Surprisingly, MSE proves to be the best metric for discriminating $\x^{76}$ and $\f^{55/76}$. This result should however be mitigated by the following observations:

\begin{itemize}
    \item metric M-LLS performs very close to MSE and even outperformed it on certain runs;
    \item the high variability in performance of MSE on different types of fakes makes it highly unreliable for authentication, as shown by its average score.
\end{itemize}

\subsection{Model stability}

Another question that we investigated is the stability of Algorithm \ref{predictor_algo} with respect to the size of the training set. We already discussed, in Section \ref{ssec:algo_param_choice}, the fact that every neighbourhood appears $100$ times on average in each template. Thus, it makes sense to run the algorithm on very small training sets. In order to measure the performance of a codebook $\mathcal{C}$ learned on a training set $\{ (\t^n,\x^n) \}$, we compare it with a reference codebook $\mathcal{C}^{ref}$ learned on the whole dataset of $720$ pairs $\{ (\t^n,\x^n) \}$. We then simply compute an average $\ell_1$-distance between the predictions:
\begin{equation}
    d_1(\mathcal{C}, \mathcal{C}^{ref}) = \frac{1}{|\Omega|} \sum_{\omega \in \Omega} |P(\omega) - P^{ref}(\omega)|.
\end{equation}

\begin{figure}[htbp]
\centerline{\includegraphics[scale=.6, trim={0cm, 0cm, 1.cm, 1.cm}]{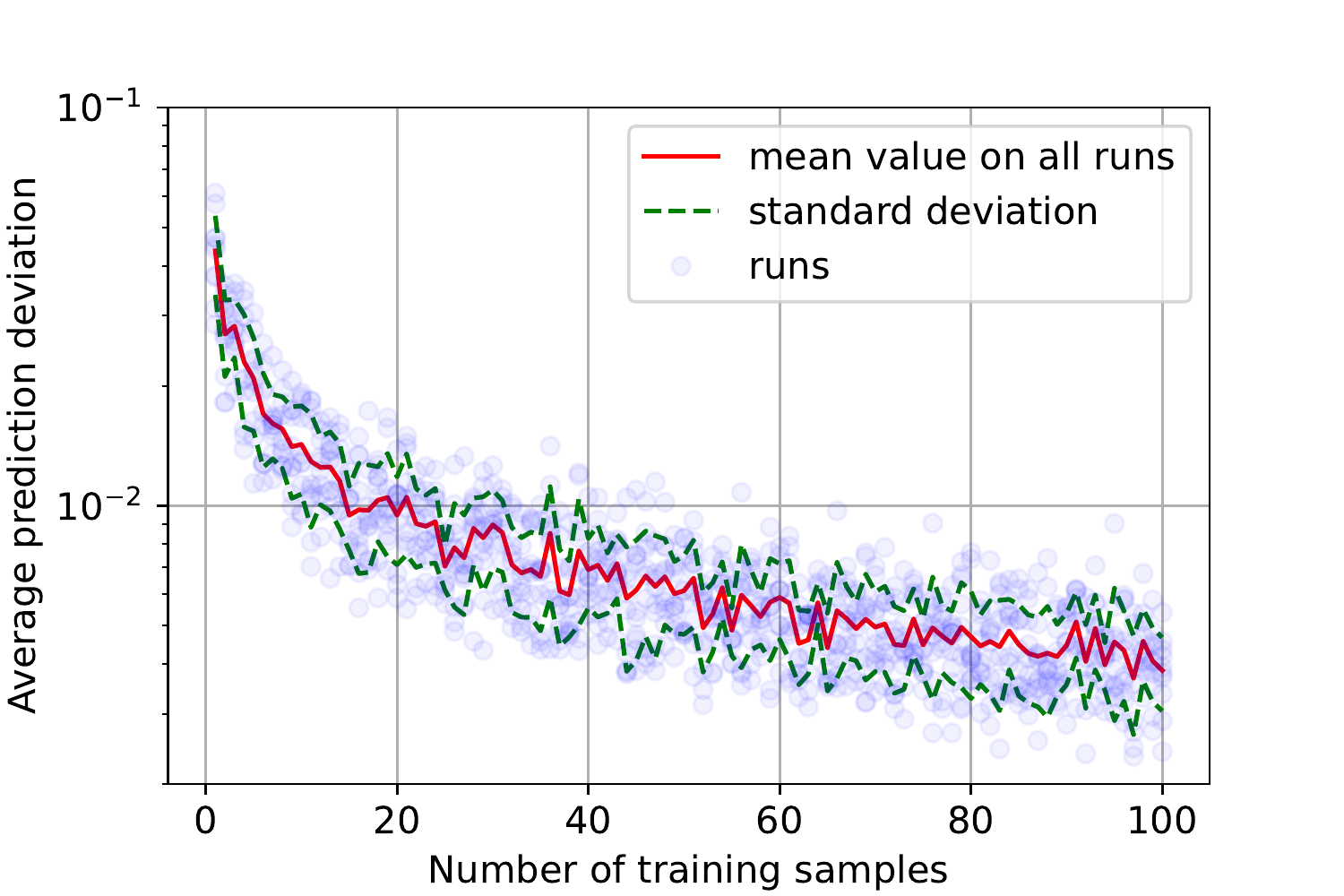}}
\caption{Variability of codebook $\mathcal{C}$ with respect to the size of the training set.}
\label{fig:train_set_dependency}
\end{figure}

Fig. \ref{fig:train_set_dependency} shows the results of this study for different training sets size with a number of samples going from $1$ to $100$. What we can see is that when using $50$ samples, the probabilities in the codebook $\mathcal{C}$ differ with the reference by less than $1 \%$ on average and the variability is very small. This explains why we decided to use $50$ training samples in our experiments.

\section{Conclusion}
\label{section:conclusion}

In this paper, we introduced a new mathematical model for the description of the Printing-Imaging channel based on local statistics.

We proposed two novel OC-authentication schemes based on this model which outperform the standard metrics used nowadays, while still maintaining full interpretability of the results. We showed that even ML-based attacks cannot fool our new authentication system. In constrast with modern deep learning approaches, our model requires very few training data and does not require much time to be run in practice, while still offering great performances against powerful ML attacks.

For future work, we aim at continuing to explore this model as the information-theoretic aspects can be deeper investigated. We also plan to replace the simple estimator with more sophisticated techniques based on neural networks and perform the comparison of the proposed approach with deep classifiers. Finally, we plan to extend the results on a new dataset acquired by several types of mobile phones which will bring more variability and new challenges for the PI channel model.

\bibliographystyle{IEEEtran}
\bibliography{bibliography.bib}

\end{document}